\documentclass[conference]{IEEEtran}
\IEEEoverridecommandlockouts
\usepackage{cite}
\usepackage{placeins}
\usepackage{amsmath,amssymb,amsfonts}
\usepackage{algorithmic}
\usepackage{graphicx}
\usepackage{textcomp}
\usepackage{caption}
\usepackage{xcolor}
\usepackage{subcaption}
\usepackage{booktabs}

\def\BibTeX{{\rm B\kern-.05em{\sc i\kern-.025em b}\kern-.08em
    T\kern-.1667em\lower.7ex\hbox{E}\kern-.125emX}}
\begin{document}

\title{Modeling PWM-Time-SOC Interaction in a Simulated Robot\\
}

\author{\IEEEauthorblockN{Vidyut Pradeep}
\textit{South Brunswick High School}\\
Monmouth Junction, New Jersey, United States\\
vidyutpradeep12@gmail.com
\and
\IEEEauthorblockN{Shirantha Welikala}
\textit{Stevens Institute of Technology}\\
Hoboken, New Jersey, United States\\
swelikal@stevens.edu}

\maketitle

\begin{abstract}
Accurate prediction of battery state of charge is needed for autonomous robots to plan movements without using up all available power. This work develops a physics and data-informed model from a simulation that predicts SOC depletion as a function of time and PWM duty cycle for a simulated 4-wheel Arduino robot. A forward-motion simulation incorporating motor electrical characteristics (resistance, inductance, back-EMF, torque constant) and mechanical dynamics (mass, drag, rolling resistance, wheel radius) was used to generate SOC time-series data across PWM values from 1-100\%. Sparse Identification of Nonlinear Dynamics (SINDy), combined with least-squares regression, was applied to construct a unified nonlinear model that captures $\text{SOC}(t, p)$. The framework allows for energy-aware planning for similar robots and can be extended to incorporate arbitrary initial SOC levels and environment-dependent parameters for real-world deployment.
\end{abstract}

\begin{IEEEkeywords}
State of charge prediction, autonomous robots, SINDy, physics-informed modeling, energy-aware planning, battery management systems
\end{IEEEkeywords}

\section{Introduction}

Energy management is a critical challenge in autonomous robotic systems. The title of this work, "Modeling PWM-Time-SOC Interaction in a Simulated Robot," refers to the relationship between Pulse Width Modulation (PWM) motor control signals, operational time, and battery State of Charge (SOC) in four-wheeled mobile robots. Understanding this interaction enables robots to predict their remaining operational capacity and avoid battery depletion during planned actions.

Autonomous mobile robots increasingly operate in environments where recharging opportunities are limited or unavailable. Without accurate energy consumption models, robots risk depleting their batteries before completing assigned tasks, leading to failures mid-action. Traditional approaches either rely on simplified linear models that fail to capture the nonlinear dynamics of the battery discharge or require extensive pre-programmed tables for every possible motion profile. Neither approach generalizes well across different operating conditions or robot configurations.

The fundamental problem addressed in this work is determining the battery state of charge during a robot's motion as a function of its control inputs. For wheeled mobile robots, the primary control input is the PWM duty cycle applied to the motors, which directly influences motor torque, current draw, and consequently, battery depletion rate. The relationship between PWM settings, time, and SOC is inherently nonlinear due to motor electrical characteristics, mechanical dynamics, and battery internal resistance effects. 

This work focuses on the forward motion of a simulated 4-wheel Arduino-based robot, developing a mathematical model that accepts time and PWM duty cycle as inputs and outputs the battery SOC at any given moment. The model enables robots to predict their energy consumption for planned movements, facilitating energy-aware path planning and task scheduling. By incorporating physics-based simulation with data-driven identification techniques, we develop interpretable governing equations that capture the essential dynamics while maintaining computational efficiency suitable for onboard implementation.

\subsection{Contributions}

This paper makes three primary contributions to energy-aware robotics. First, we develop a unified nonlinear model for battery SOC prediction that explicitly captures the coupled effects of PWM duty cycle and time using Sparse Identification of Nonlinear Dynamics (SINDy). Unlike lookup-table approaches that require extensive pre-recording of motion profiles, our analytical model generalizes across the full range of PWM values and can compute SOC for arbitrary time horizons. The model includes polynomial terms up to third order in PWM and logarithmic terms in time, with specific functional forms $\text{SOC}(t,p)$, $\text{SOC}(p)$, and $\frac{\partial \text{SOC}}{\partial t}$ derived from first principles and validated against simulation data.

Second, we demonstrate a systematic methodology for physics-informed, data-driven modeling of robotic energy consumption that bridges simulation and real-world deployment. Our approach incorporates detailed motor electrical models (resistance, inductance, back-EMF, torque constant) and mechanical dynamics (mass, drag, rolling resistance) in simulation to generate training data, then applies sparse regression to identify governing equations. This methodology differs from purely empirical approaches by grounding the candidate function library in physical understanding, and from purely theoretical approaches by learning model coefficients directly from observed behavior. The resulting models achieve prediction errors averaging 0.162\% with maximum errors under 0.82\% across validation datasets. 

Third, we provide a complete framework for extending this modeling approach to real-world deployment with varying initial conditions and environmental parameters. While the current implementation assumes fully charged initial states, we outline the path toward $\text{SOC}(t, p, \text{SOC}_0)$ formulations and discuss integration of environment-dependent parameters such as terrain-specific rolling resistance. This flexibility enables adaptation to different operating conditions without complete model retraining.

The remainder of this paper is organized as follows. Section II reviews related work in robot energy modeling and energy-efficient planning. Section III introduces the Sparse Identification of Nonlinear Dynamics framework used for model discovery. Section IV provides background on battery State of Charge for mobile robots, including fundamental governing equations. Section V details our simulation environment, including motor models, mechanical dynamics, and data generation procedures. Section VI describes the methodology and data fitting process. Section VII presents the derived SOC models with numerical coefficients and validation results. Section VIII discusses model accuracy, limitations, and future extensions. Section IX concludes the paper.

\section{Literature Review}

Energy-efficient robot scheduling has been addressed through dynamic programming approaches that optimize trajectories by scaling execution times along predefined paths. In \cite{b4}, a method for energy-efficient robot scheduling is presented that uses dynamic programming to generate energy-optimal trajectories by dynamically scaling execution times along predefined paths. Their dynamic time scaling approach achieved 18-28\% energy reduction compared to time-optimal execution in a four-robot case study, significantly outperforming linear scaling methods. However, the approach requires predefined geometric paths from existing trajectories and relies on simplified motor models that don't account for regenerative braking. Additionally, the method assumes operations start from known states and doesn't use arbitrary initial conditions. In contrast, our work requires no predefined path geometry and learns a closed-form $\text{SOC}(t,p)$ model directly from motor dynamics, enabling energy prediction for arbitrary PWM inputs without prior trajectory knowledge.

Empirical characterization of robot power consumption has provided valuable insights into component-level energy usage. An empirical study of the Pioneer 3DX mobile robot in \cite{b5} measured power consumption across all major components and built linear power models based on experimental data. Their comprehensive power breakdown revealed that motion accounts for less than 50\% of total power consumption, demonstrating the importance of considering all components in energy-efficient designs, and their proposed techniques showed potential savings of 11\%-44\% in specific subsystems. However, the study only built empirical models based on measured data from one specific robot platform without deriving generalizable mathematical relationships from underlying physics. In contrast, our approach derives candidate functions from motor and battery physics, producing a generalizable analytical model that is not tied to any single platform and can be adapted by substituting robot-specific parameters.

Physics-based simulation has been used to create realistic models of robot actuators and power systems. A realistic servomotor model for humanoid robot simulation incorporating DC motor electrical characteristics, friction models, and closed-loop controller dynamics was created in \cite{b6}, validating their simulator by comparing behaviors between simulated and real robots. A physics-based approach captures realistic servomotor behavior through parameter identification from real measurements, and validation showed close agreement during walking movements. However, their controller model does not assume frictional variations during operation. In contrast, our work uses physics-based simulation not as an end goal but as a data source for SINDy-based model discovery, yielding compact closed-form SOC equations rather than high-fidelity but computationally expensive simulation replicas.

Alternative sensing modalities have been explored for robot state estimation that could inform energy models. Estimation methods for mobile robots using optical flow sensors have been proposed \cite{b7}, including both a single-sensor method relying on kinematic constraints and a dual-sensor method that eliminates these constraints, with optimal sensor placement to minimize measurement errors. Their dual-sensor rigid-body method maintained accurate position estimates even when wheel slip and kinematic violations occurred. However, the optical flow sensors require precise ground clearance to function properly, limiting terrain adaptability and sometimes requiring sensors to be pressed against the ground, which increases friction. In contrast, our model requires no additional sensing hardware beyond what is already used for PWM motor control, making it immediately deployable on existing robot platforms without sensor modifications.

Energy-efficient path planning has been addressed by modifying classical planning algorithms to incorporate energy models. An energy-efficient path planning approach \cite{b8} integrates modified cost and heuristic functions into the A* algorithm, building a comprehensive energy model for a three-wheeled omnidirectional mobile robot, including motor losses, kinetic energy losses, friction losses, and electronic component consumption. Their modified A* algorithm with angular penalty functions reduced path turns and simulation time while maintaining safe distances from obstacles, achieving 1.17-16.84\% energy savings compared to existing methods. However, the method requires significant preprocessing time to compute heuristic values, which may limit real-time replanning capabilities in dynamic environments. In contrast, our closed-form $\text{SOC}(t,p)$ model evaluates in microseconds and can serve as a lightweight energy cost function within any planner, including A*, without requiring separate preprocessing or lookup tables.

Terrain-aware planning considers how surface properties affect energy consumption at different speeds. Applying an energy-efficient path planning approach for low-speed autonomous electric vehicles using a modified Dijkstra algorithm that considers rolling resistance variations across different terrain types, building energy-based and distance-based adjacency matrices from rolling resistance maps and vehicle dynamics, is outlined in \cite{b9}. Their approach achieved 13-46\% energy savings compared to traditional distance-based planning across multiple scenarios while accepting reasonable path length increases, and explicitly models how rolling resistance dominates energy consumption at low speeds. However, the approach is limited to offline global planning with static maps and cannot handle dynamic obstacles or changing terrain conditions without complete re-planning. In contrast, our model encapsulates terrain effects through the rolling resistance parameter $C_{rr}$, which can be updated at runtime for different surfaces without retraining, enabling online adaptation to changing terrain conditions.

Marine robotics has also benefited from energy-aware planning that accounts for environmental forces. In \cite{b10}, a Voronoi-Visibility energy-efficient path planning algorithm for USVs combines Voronoi diagrams, Visibility graphs, and Dijkstra's search with an energy consumption model considering sea current effects, generating collision-free paths while maintaining configurable safety distances from coastlines. Their approach achieved 1.27-52.84\% energy savings compared to shortest-path planning across multiple missions while maintaining computational efficiency. However, the approach assumes static sea current conditions during missions and does not allow USVs to change speed adaptively based on current states, limiting optimization potential for long-range missions. In contrast, our model explicitly captures time-varying SOC dynamics through its temporal terms, enabling a robot to adapt its PWM setpoint mid-mission based on continuously updated SOC predictions rather than relying on static pre-planned energy budgets.

Communication energy consumption in networked mobile robots has been addressed alongside motion energy. Ooi and Schindelhauer \cite{b11} developed the first approach to jointly optimize energy consumption for both communication and motion in wireless robots, proposing position-critical and constant bit-rate communication models with approximation algorithms based on Dijkstra's shortest path that account for polynomial communication energy increases with distance. Their approach achieved up to 48.71\% energy savings compared to straight-line movement, demonstrating that communication costs can dominate, especially at high data rates, and the framework recognizes that shortest paths are not energy-optimal when wireless costs are considered. However, their model was only proportional to distance, and the actual computational complexity was noted to be too high for practical applications. In contrast, our model focuses on the dominant motion energy component and provides an analytically tractable, low-complexity expression that is practical for onboard real-time computation, and could be extended to incorporate communication costs as an additive term.

Energy consumption analysis specifically for robotic task execution has established methodologies for measuring and characterizing power usage. In \cite{b1}, an energy consumption analysis procedure for robotic applications examines different task motions, providing insights into how various movements affect overall energy usage. This work emphasizes the need for motion-specific energy models that can inform task planning and execution strategies. In contrast, our work goes beyond characterization by producing a predictive analytical model — rather than measuring energy after the fact, our $\text{SOC}(t,p)$ formulation allows a robot to forecast its remaining charge before executing a motion, enabling proactive energy-aware decision making.

Our work differs from these approaches by developing a unified analytical model that directly captures the nonlinear relationship between control inputs (PWM), time, and battery SOC through physics-informed sparse regression. Rather than relying on extensive pre-recorded motion libraries or requiring predefined paths, our model provides closed-form equations that generalize across operating conditions. Unlike purely empirical studies, we ground our approach in physically meaningful candidate functions derived from motor and battery dynamics, yielding interpretable models suitable for real-time onboard computation.

\section{Preliminaries: Sparse Identification of Nonlinear Dynamics (SINDy)}

Modeling dynamical systems often requires discovering governing equations that describe how a state evolves over time. Traditional approaches depend on either first-principles derivations or black-box models, both of which come with limitations: first-principles models may be incomplete or require extensive domain expertise, while black-box models lack interpretability and physical insight into the underlying dynamics. Sparse Identification of Nonlinear Dynamics (SINDy) offers an alternative by directly learning differential equations from data while enforcing sparsity, resulting in interpretable models grounded in the system's physics.

\subsection{The SINDy Framework}

Consider a dynamical system where measurements of the state $\mathbf{x}(t) \in \mathbb{R}^n$ are available at discrete time points. The goal is to discover a governing equation of the form:
\begin{equation}
\frac{d\mathbf{x}}{dt} = \mathbf{f}(\mathbf{x}),
\end{equation}
where $\mathbf{f}$ represents the unknown dynamics. SINDy assumes that the right-hand side can be represented as a sparse linear combination of candidate functions from a predefined library.

First, we construct a data matrix $\mathbf{X}$ containing state measurements at different time points as
\begin{equation}
\mathbf{X} = \begin{bmatrix}
\mathbf{x}^T(t_1) \\
\mathbf{x}^T(t_2) \\
\vdots \\
\mathbf{x}^T(t_m)
\end{bmatrix} = \begin{bmatrix}
x_1(t_1) & x_2(t_1) & \cdots & x_n(t_1) \\
x_1(t_2) & x_2(t_2) & \cdots & x_n(t_2) \\
\vdots & \vdots & \ddots & \vdots \\
x_1(t_m) & x_2(t_m) & \cdots & x_n(t_m)
\end{bmatrix}.
\end{equation}

Next, we compute or approximate the time derivatives to form the derivative matrix as
\begin{equation}
\dot{\mathbf{X}} = \begin{bmatrix}
\dot{\mathbf{x}}^T(t_1) \\
\dot{\mathbf{x}}^T(t_2) \\
\vdots \\
\dot{\mathbf{x}}^T(t_m)
\end{bmatrix}.
\end{equation}

The core of SINDy is constructing a library matrix $\boldsymbol{\Theta}(\mathbf{X})$ containing candidate functions evaluated at each measurement point:
\begin{equation}
\boldsymbol{\Theta}(\mathbf{X}) = \begin{bmatrix}
1 & \mathbf{X} & \mathbf{X}^2 & \cdots & \sin(\mathbf{X}) & \cdots
\end{bmatrix}.
\end{equation}

Each column represents a candidate function (e.g., constant, linear terms, polynomials, trigonometric functions, logarithmic functions) evaluated at all time points. The library is designed based on physical intuition about what functions might govern the system.

The governing equation can then be expressed as a sparse regression problem such as
\begin{equation}
\dot{\mathbf{X}} = \boldsymbol{\Theta}(\mathbf{X}) \boldsymbol{\Xi},
\end{equation}
where $\boldsymbol{\Xi}$ is a sparse coefficient matrix. Each column of $\boldsymbol{\Xi}$ contains the coefficients for one state variable's dynamics.

\subsection{Sparse Regression}

To identify which candidate functions are truly necessary, SINDy employs sparse regression techniques. The goal is to solve a \emph{regularized least-squares} problem. Specifically, for a matrix $\mathbf{A} \in \mathbb{R}^{m \times n}$ and a vector $\mathbf{b} \in \mathbb{R}^m$, the \textbf{2-norm} (Euclidean norm) of a vector $\mathbf{v} \in \mathbb{R}^k$ is defined as
\[
\|\mathbf{v}\|_2 = \sqrt{\sum_{i=1}^{k} v_i^2},
\]
and the \textbf{1-norm} (sparsity-inducing norm) is defined as
\[
\|\mathbf{v}\|_1 = \sum_{i=1}^{k} |v_i|.
\]
An alternative sparsity-promoting approach replaces the 2-norm penalty 
with a 1-norm penalty, commonly known as the Least Absolute Shrinkage 
and Selection Operator (LASSO). The regularized objective becomes
\begin{equation}
J(\boldsymbol{\Xi}) = \|\dot{\mathbf{X}} - \boldsymbol{\Theta}(\mathbf{X})\boldsymbol{\Xi}\|_2^2 + \lambda\|\boldsymbol{\Xi}\|_1,
\end{equation}
where $\|\boldsymbol{\Xi}\|_1 = \sum_{i,j} |\xi_{ij}|$ is the entry-wise 
sum of absolute values defined above. Unlike Ridge regression, the 1-norm 
penalty drives coefficients to \emph{exactly} zero rather than merely 
shrinking them, making it a natural fit for SINDy where true sparsity in 
$\boldsymbol{\Xi}$ is desired. In the context of PWM-SOC dynamics, 
1-norm SINDy would automatically select the minimal subset of candidate 
functions from $\boldsymbol{\Theta}$ — such as $\ln(1+t)$, $p^2$, and 
$\frac{t}{p+1}$ — without requiring the sequential thresholding step. 
However, because LASSO lacks a closed-form solution and requires iterative 
solvers such as coordinate descent, it carries higher computational cost 
than the approach adopted here.
\par These norms apply to vectors; when applied to a matrix $\boldsymbol{\Xi}$, $\|\boldsymbol{\Xi}\|_F$ denotes the Frobenius norm (entry-wise 2-norm), while $\|\boldsymbol{\Xi}\|_1$ is understood as the entry-wise sum of absolute values. The regularized least-squares objective is then
\begin{equation}
J(\boldsymbol{\Xi}) = \|\dot{\mathbf{X}} - \boldsymbol{\Theta}(\mathbf{X})\boldsymbol{\Xi}\|_2^2 + \lambda \|\boldsymbol{\Xi}\|_2^2,
\end{equation}
where the first term (using the squared 2-norm) enforces data accuracy and the second term (the 2-norm, with regularization parameter $\lambda > 0$) penalizes large coefficients, shrinking them toward zero. This formulation is known as \emph{Ridge regression} (or Tikhonov regularization) and has a closed-form solution:
\[
\boldsymbol{\Xi} = (\boldsymbol{\Theta}^T\boldsymbol{\Theta} + \lambda \mathbf{I})^{-1}\boldsymbol{\Theta}^T\dot{\mathbf{X}}.
\]
Unlike the 1-norm penalty, the 2-norm penalty shrinks all coefficients toward zero but does not produce exact zeros; sequential thresholding (described below) is therefore used in addition to enforce sparsity.

Alternatively, sequential thresholding can be used:
\begin{enumerate}
\item Solve the least-squares problem: $\boldsymbol{\Xi} = (\boldsymbol{\Theta}^T\boldsymbol{\Theta})^{-1}\boldsymbol{\Theta}^T\dot{\mathbf{X}}$
\item Threshold small coefficients to zero (set coefficients below threshold to zero)
\item Repeat with a reduced library containing only active terms until convergence
\end{enumerate}

This process identifies a minimal set of terms that accurately describe the dynamics, yielding interpretable governing equations. The sparsity constraint is crucial because most physical systems are governed by relatively few active terms, even though many candidate functions might plausibly contribute.

\section{Battery State of Charge for Mobile Robots}

This section provides the theoretical background on battery State of Charge (SOC) as it applies to autonomous mobile robots, including the fundamental governing equations, the role of motor current and mechanical dynamics, and the three complementary SOC models that emerge from those physical relationships.

\subsection{Mobile Robot Energy Consumption}

A mobile robot draws power from its battery to operate all of its subsystems simultaneously. These include the drive motors for locomotion, the onboard microcontroller and embedded computing hardware, sensors (e.g., encoders, IMUs, cameras), and communications modules. Among these, the drive motors typically dominate power consumption during motion, particularly at high duty cycles, because motor current scales with mechanical load and the resulting $I^2 R$ losses in the motor windings and battery internal resistance are significant.

For a four-wheeled robot driven by DC motors under PWM control, the instantaneous electrical power drawn from the battery is
\begin{equation}
P_{\text{total}}(t) = V_{\text{batt}}(t) \cdot I_{\text{motor}}(t),
\end{equation}
where $V_{\text{batt}}(t)$ is the terminal voltage of the battery and $I_{\text{motor}}(t)$ is the motor current drawn by all four motors. It is worth noting that a small portion of the total motor current also supplies the onboard microcontroller and sensors; however, because these components consume far less power than the drive motors during active locomotion, $I_{\text{motor}}$ is treated here as the dominant contributor to battery depletion, with any residual processing draw handled as a constant background offset if needed.

The terminal voltage drops as current increases due to the battery's internal resistance $R_{\text{int}}$ can be expressed as
\begin{equation}
V_{\text{batt}}(t) = V_{\text{oc}} - I_{\text{motor}}(t) \cdot R_{\text{int}},
\end{equation}
where $V_{\text{oc}}$ is the open-circuit voltage. Substituting (8) into (9) shows that power scales nonlinearly with motor current:
\begin{equation}
P_{\text{total}}(t) = \bigl(V_{\text{oc}} - I_{\text{motor}}(t) \cdot R_{\text{int}}\bigr)\, I_{\text{motor}}(t).
\end{equation}
This voltage droop means that higher motor loads produce a nonlinear increase in power consumption, which directly motivates the polynomial PWM terms in the SOC models derived in Section VII.

Crucially, this electrical power does not disappear; it is converted by the motors into mechanical work that must overcome the forces acting on the robot. At steady-state forward velocity $v$, the net mechanical power balance is:
\begin{equation}
P_{\text{mech}}(t) = \bigl(F_{\text{drag}}(v) + F_{\text{roll}}(v) + F_{\text{wind}}(v)\bigr)\cdot v
\end{equation}
where $F_{\text{drag}}$ is aerodynamic drag, $F_{\text{roll}}$ is rolling resistance, and $F_{\text{wind}}$ accounts for any headwind. Because drag and rolling resistance both depend on velocity, and velocity is itself a function of PWM and time through the motor–mechanical dynamics, both $t$ and $p$ appear explicitly in $P_{\text{total}}(t)$ and therefore in $\frac{d(\text{SOC})}{dt}$. This coupling is the physical justification for requiring a two-variable model $\text{SOC}(t,p)$ rather than a simple linear expression.

Processing energy—consumed by the microcontroller and sensors—is generally much smaller than motor energy during active locomotion and is treated as a constant background draw in our model. For more comprehensive whole-system energy budgeting, this term can be added as a fixed offset to the motor-driven SOC depletion rate.

\subsection{Fundamental Governing Equations for SOC}

Battery SOC quantifies the remaining charge as a fraction of the battery's total capacity $Q_{\text{cap}}$ (in ampere-hours). The coulomb-counting definition of SOC takes the form:
\begin{equation}
\text{SOC}(t) = \text{SOC}(0) - \frac{1}{Q_{\text{cap}}} \int_0^t I(\tau)\, d\tau,
\end{equation}
where $I(\tau) \geq 0$ is the discharge current and $Q_{\text{cap}}$ is the rated capacity. Differentiating yields the governing differential equation for SOC:
\begin{equation}
\frac{d(\text{SOC})}{dt} = -\frac{I_{\text{motor}}(t)}{Q_{\text{cap}}}.
\end{equation}

The discharge current $I_{\text{motor}}(t)$ is itself a function of the motor operating point. Under PWM control with duty cycle $p \in [0, 1]$, the effective voltage applied to each motor is $V_{\text{mot}} = p \cdot V_{\text{batt}}$. The motor current dynamics (neglecting inductance for steady-state) give
\begin{equation}
I_{\text{motor}} = \frac{p \cdot V_{\text{batt}} - K_e \omega}{R},
\end{equation}
where $K_e$ is the back-EMF constant, $\omega$ is the motor angular velocity, and $R$ is the armature resistance. The angular velocity $\omega$ is directly related to the wheel's translational velocity $v$ via $\omega = v/r$ (where $r$ is the wheel radius), so that as the robot accelerates toward its steady-state speed, $I_{\text{motor}}$ decreases from its large startup value. Meanwhile, the mechanical equation of motion defined as
\begin{equation}
m\dot{v} = F_{\text{drive}} - F_{\text{drag}}(v) - F_{\text{roll}} - F_{\text{wind}},
\end{equation}
links velocity evolution to the friction and drag forces that the motors must overcome. Because $\omega$ (and hence $I_{\text{motor}}$) evolves according to these mechanical dynamics, the current $I_{\text{motor}}(t)$ and hence $\frac{d(\text{SOC})}{dt}$ are nonlinear functions of both $p$ and $t$. This nonlinearity motivates the data-driven SINDy approach: rather than solving the fully coupled differential equations analytically, we learn the functional form of $\text{SOC}(t, p)$ directly from simulation data.

\subsection{Theoretical SOC Model Derivation}
\label{sec:theoretical_soc}
Using the governing equations established above, we can derive a 
physically motivated closed-form approximation for $\text{SOC}(t,p)$ 
by analyzing two distinct operating regimes. These are physically 
motivated approximations rather than exact solutions, as the fully 
coupled system of equations does not admit a closed form; their 
value lies in predicting the functional structure that SINDy 
later recovers from data.

The variables appearing throughout this derivation are defined as 
follows: $t \geq 0$ is time in seconds; $p \in [0,1]$ is the PWM 
duty cycle fraction; $V_{\text{oc}}$ is the open-circuit (no-load) 
battery voltage in volts; $R$ is the DC armature resistance of one 
motor in ohms; $R_{\text{int}}$ is the battery internal resistance 
in ohms; $K_e$ is the back-EMF constant in V$\cdot$s/rad; $K_t$ is 
the motor torque constant in N$\cdot$m/A (equal to $K_e$ for an 
ideal DC motor); $\omega$ is the motor shaft angular velocity in 
rad/s; $v$ is the robot's translational velocity in m/s; $r$ is 
the wheel radius in m, linking $\omega$ and $v$ via $\omega = v/r$; 
$m$ is the total robot mass in kg; $g = 9.81\ \text{m/s}^2$ is 
gravitational acceleration; $C_{rr}$ is the dimensionless rolling 
resistance coefficient; $\rho$ is the air density in kg/m$^3$; 
$C_d$ is the dimensionless aerodynamic drag coefficient; $A$ is 
the robot's frontal cross-sectional area in m$^2$; $v_w$ is the 
ambient headwind speed in m/s; $Q_{\text{cap}}$ is the battery 
capacity in ampere-seconds (A$\cdot$s), obtained by multiplying 
the rated mAh value by 3.6; and $I_{\text{motor}}(t)$ is the 
total motor current drawn from the battery at time $t$ in amperes.

\subsubsection{Transient Regime}

At startup, the robot is at rest, so $v \approx 0$ and 
$\omega \approx 0$. Substituting the battery terminal voltage 
from~(9) into the motor current expression~(14) and solving 
for $I_{\text{motor}}$:
\begin{equation}
I_{\text{motor}}(R + pR_{\text{int}}) = pV_{\text{oc}} - K_e\omega.
\end{equation}
Setting $\omega = 0$ yields the peak startup current:
\begin{equation}
I_{\text{motor}}\big|_{t \to 0} \approx \frac{pV_{\text{oc}}}{R + pR_{\text{int}}}.
\label{eq:I_transient}
\end{equation}

In the transient regime the motor has not yet accelerated, so 
the back-EMF $K_e\omega$ is negligible and the current is 
dominated by the resistive voltage divider between $R$ and 
$pR_{\text{int}}$. As the robot accelerates, $\omega$ grows 
and back-EMF increasingly opposes the applied voltage, causing 
$I_{\text{motor}}$ to decay. This decay is well approximated 
by an exponential:
\begin{equation}
I_{\text{motor}}^{\text{trans}}(t) \approx 
\frac{pV_{\text{oc}}}{R + pR_{\text{int}}}\, e^{-\alpha t},
\label{eq:I_trans_exp}
\end{equation}
where $\alpha > 0$ is a lumped time constant that depends on 
the motor electrical and mechanical parameters. The 
corresponding transient contribution to SOC depletion is 
found by integrating~\eqref{eq:I_trans_exp} via 
coulomb-counting~(12):
\begin{align}
\Delta\text{SOC}_{\text{trans}}(t,p) 
&= \frac{1}{Q_{\text{cap}}} \int_0^t 
   I_{\text{motor}}^{\text{trans}}(\tau)\,d\tau \notag\\
&= \frac{1}{Q_{\text{cap}}} \int_0^t 
   \frac{pV_{\text{oc}}}{R + pR_{\text{int}}}\,e^{-\alpha\tau}\,d\tau \notag\\
&= \frac{pV_{\text{oc}}}{Q_{\text{cap}}(R + pR_{\text{int}})}
   \left[-\frac{1}{\alpha}e^{-\alpha\tau}\right]_0^t \notag\\
&= \frac{pV_{\text{oc}}}{\alpha\, Q_{\text{cap}}(R + pR_{\text{int}})}
   \bigl(1 - e^{-\alpha t}\bigr).
\label{eq:deltaSOC_trans_exact}
\end{align}
For moderate $t$ the factor $(1 - e^{-\alpha t})$ is 
well approximated by $\ln(1+t)$ (both are zero at $t=0$, 
grow rapidly for small $t$, and saturate for large $t$), 
so absorbing $\alpha$ into the overall coefficient gives:
\begin{equation}
\Delta\text{SOC}_{\text{trans}}(t,p) \approx 
\frac{pV_{\text{oc}}}{Q_{\text{cap}}(R + pR_{\text{int}})}\,\ln(1+t).
\label{eq:deltaSOC_trans}
\end{equation}

\subsubsection{Steady-State Regime}

Once the robot reaches constant velocity ($\dot{v} = 0$), 
setting the left-hand side of~(15) to zero gives the 
force balance:
\begin{equation}
\frac{K_t}{r} \cdot I_{\text{motor}}^{ss} = C_{rr}mg 
+ \frac{1}{2}\rho C_d A\bigl(v_{ss} + v_w\bigr)^2,
\label{eq:force_balance}
\end{equation}
where the left-hand side is the net drive force produced by 
the four motors (torque $K_t I_{\text{motor}}^{ss}$ divided 
by wheel radius $r$), and the right-hand side contains two 
resistive forces: the rolling resistance $C_{rr}mg$ 
(proportional to the normal force) and the aerodynamic drag 
$\tfrac{1}{2}\rho C_d A(v_{ss}+v_w)^2$ (quadratic in the 
sum of robot velocity and headwind). Solving~\eqref{eq:force_balance} 
for the steady-state current:
\begin{equation}
I_{\text{motor}}^{ss}(p) = \frac{r}{K_t}
\left(C_{rr}mg + \frac{1}{2}\rho C_d A\bigl(v_{ss}(p) 
+ v_w\bigr)^2\right),
\label{eq:I_ss}
\end{equation}
where $v_{ss}(p)$ is the steady-state velocity at duty cycle 
$p$. Because $I_{\text{motor}}^{ss}$ is approximately constant 
in this regime, the SOC depletion rate is also approximately 
constant, and the steady-state contribution to SOC depletion 
is found by integrating the constant current over time:
\begin{align}
\Delta\text{SOC}_{\text{ss}}(t,p) 
&= \frac{1}{Q_{\text{cap}}} \int_0^t 
   I_{\text{motor}}^{ss}(p)\,d\tau \notag\\
&= \frac{I_{\text{motor}}^{ss}(p)}{Q_{\text{cap}}}
   \int_0^t d\tau \notag\\
&= \frac{I_{\text{motor}}^{ss}(p)}{Q_{\text{cap}}}\cdot t,
\label{eq:deltaSOC_ss}
\end{align}
producing the linear-in-$t$ term that dominates at large $t$.

\subsubsection{Bridging the Two Regimes}

The full SOC trajectory is approximated by starting from a 
fully charged battery ($\text{SOC}=100$) and subtracting 
both contributions derived in~\eqref{eq:deltaSOC_trans} 
and~\eqref{eq:deltaSOC_ss}:
\begin{align}
\text{SOC}(t,p) 
&= 100 - \Delta\text{SOC}_{\text{trans}}(t,p) 
        - \Delta\text{SOC}_{\text{ss}}(t,p) \notag\\
&= 100 
   - \frac{1}{Q_{\text{cap}}} \int_0^t 
     I_{\text{motor}}^{\text{trans}}(\tau)\,d\tau
   - \frac{1}{Q_{\text{cap}}} \int_0^t 
     I_{\text{motor}}^{ss}(p)\,d\tau \notag\\
&\approx 100 
- \underbrace{\frac{pV_{\text{oc}}}{Q_{\text{cap}}(R + pR_{\text{int}})}}_{\text{transient coefficient}(p)}\ln(1+t)
- \underbrace{\frac{I_{\text{motor}}^{ss}(p)}{Q_{\text{cap}}}}_{\text{steady-state rate}(p)} \cdot t.
\label{eq:SOC_composite}
\end{align}

\subsection{Why a Coupled $(t,p)$ Model Is Necessary}
\label{sec:why_coupled}

A simple approach might assume that SOC depletion is linear in both time and PWM—i.e., that a robot at 90\% PWM simply depletes its battery at a constant rate proportional to $p$. However, three physical effects conspire to make this assumption incorrect:

\begin{enumerate}
\item \textbf{Velocity-dependent friction and drag.} From (14), the resistive forces $F_{\text{drag}}(v)$ and $F_{\text{roll}}$ depend on the instantaneous velocity. As the robot accelerates after startup, these forces increase, changing the required motor torque and therefore $I_{\text{motor}}$ over time—even at a fixed $p$. This means the depletion rate $\frac{d(\text{SOC})}{dt}$ is not constant in time; it has a pronounced transient period followed by a slower, approximately linear regime.

\item \textbf{Battery internal resistance.} As shown in (9), $P_{\text{total}}$ depends quadratically on $I_{\text{motor}}$, so the initial current surge at motor startup causes a disproportionately steep drop in SOC. This ``hook'' at $t=0$ (visible in Figures~\ref{fig:90pwm} and~\ref{fig:40pwm}) cannot be captured by a linear-in-time model; it requires terms such as $\ln(1+t)$ that are large near $t=0$ and diminish as the system settles.

\item \textbf{Nonlinear PWM-to-current mapping.} The quadratic and cubic PWM terms in all three models reflect the fact that doubling $p$ more than doubles $I_{\text{motor}}$, due to the nonlinear back-EMF and the quadratic $I^2R$ loss structure.
\end{enumerate}

Together, these effects demand a model with explicit time dependence and higher-order PWM terms—precisely the structure discovered by SINDy and presented next.

\section{Simulation Framework}

Figure 1 presents the complete simulation workflow. The process begins with two parallel inputs: the PWM signal and the physical motor dimensions. These feed into the Electric Model, which computes back-EMF and motor current, while the physical dimensions inform the Torque Block. The torque calculations then incorporate terrain conditions, gear ratio, and required torque to drive the Vehicle Dynamics subsystem. In parallel, the PWM Block computes power losses, which flow into the Battery Current Draw. The current draw is combined with thermal models and battery temperature alongside physical battery parameters to construct the Battery Model. The Battery Model then produces the final SOC Update output.

This section describes the physics-based simulation environment and the application of SINDy for model identification.

\begin{figure}[!t]
\centering
\includegraphics[width=\columnwidth]{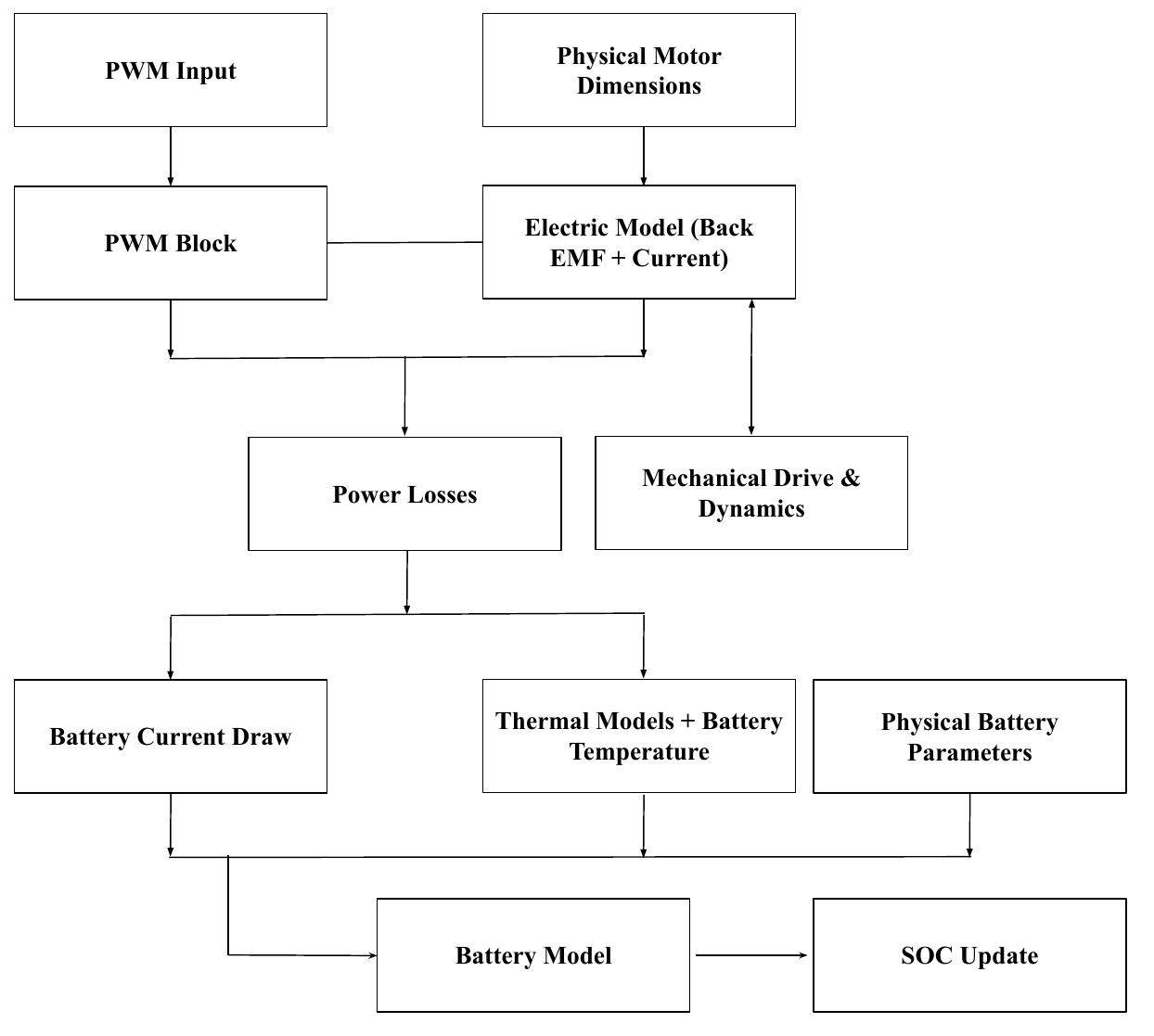}
\caption{Simulation workflow showing the progression of the data collection from the MATLAB simulated robot. Illustrates the pathways the code goes to before giving an output.}
\label{fig:Simulation Block Diagram}
\end{figure}

\begin{figure}[!t]
\centering
\includegraphics[width=\columnwidth]{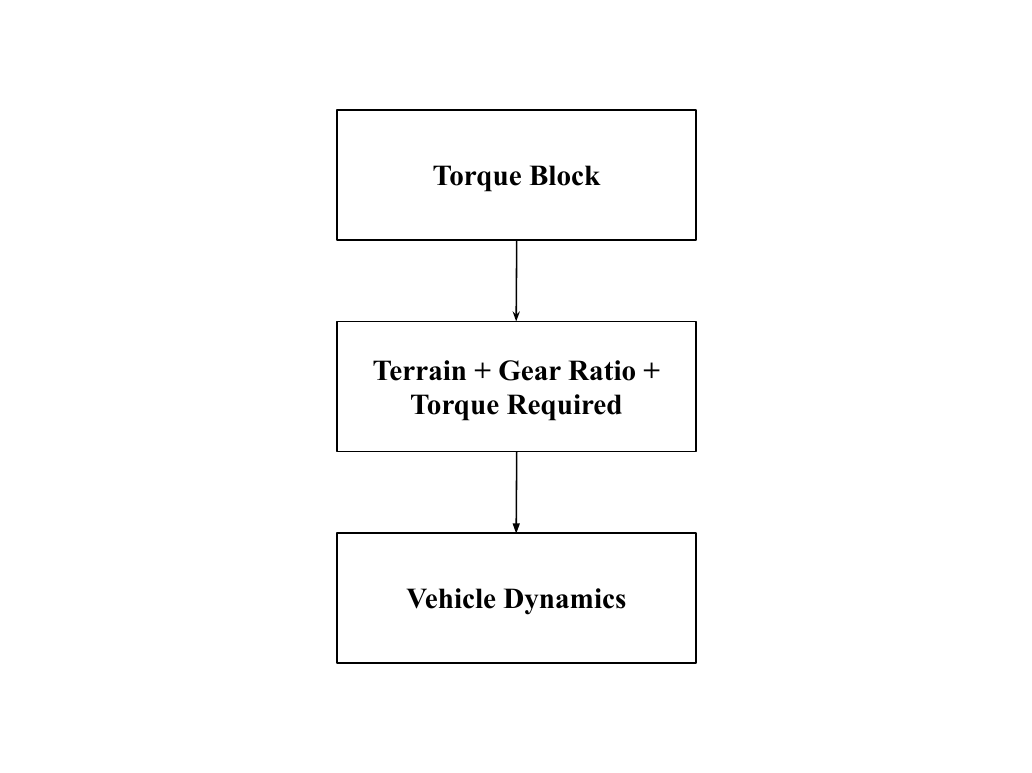}
\caption{Expanded block of the Mechanical Drive and Dynamics block}
\label{fig:Simulation Block Diagram}
\end{figure}

\subsubsection{Simulation Parameters}

The simulation uses representative values for a medium-scale Arduino-controlled wheeled robotic platform. The parameters are grouped into motor, robot, and battery subsystems as follows:

\begin{itemize}
\item \textbf{Motor:} $R = 0.5\,\Omega$, $L = 2\,\text{mH}$, $K_t = K_e = 0.02\,\text{Nm/A}$, motor efficiency $\eta = 0.80$
\item \textbf{Robot:} $m = 5\,\text{kg}$, wheel radius $r = 0.1\,\text{m}$, drag coefficient $C_d = 1.2$, rolling resistance coefficient $C_{rr} = 0.02$
\item \textbf{Battery:} Capacity $Q = 2500\,\text{mAh}$, nominal voltage $V = 12\,\text{V}$, internal resistance $R_{int} = 0.05\,\Omega$
\end{itemize}

Additional environmental and operational parameters are included to improve physical realism. These include air density $\rho = 1.225\,\text{kg/m}^3$, frontal area $A = 0.05\,\text{m}^2$, wind speed $v_w = 1\,\text{m/s}$, and gravitational acceleration $g = 9.81\,\text{m/s}^2$.

These values are representative of hobby-grade autonomous mobile robots. For deployment on physical robotic systems, these parameters can be refined through manufacturer specifications or experimental system identification. The simulation framework is modular and allows straightforward parameter modification for different robotic platforms.

\section{Methodology}

\begin{figure}[!t]
\centering
\includegraphics[width=\columnwidth]{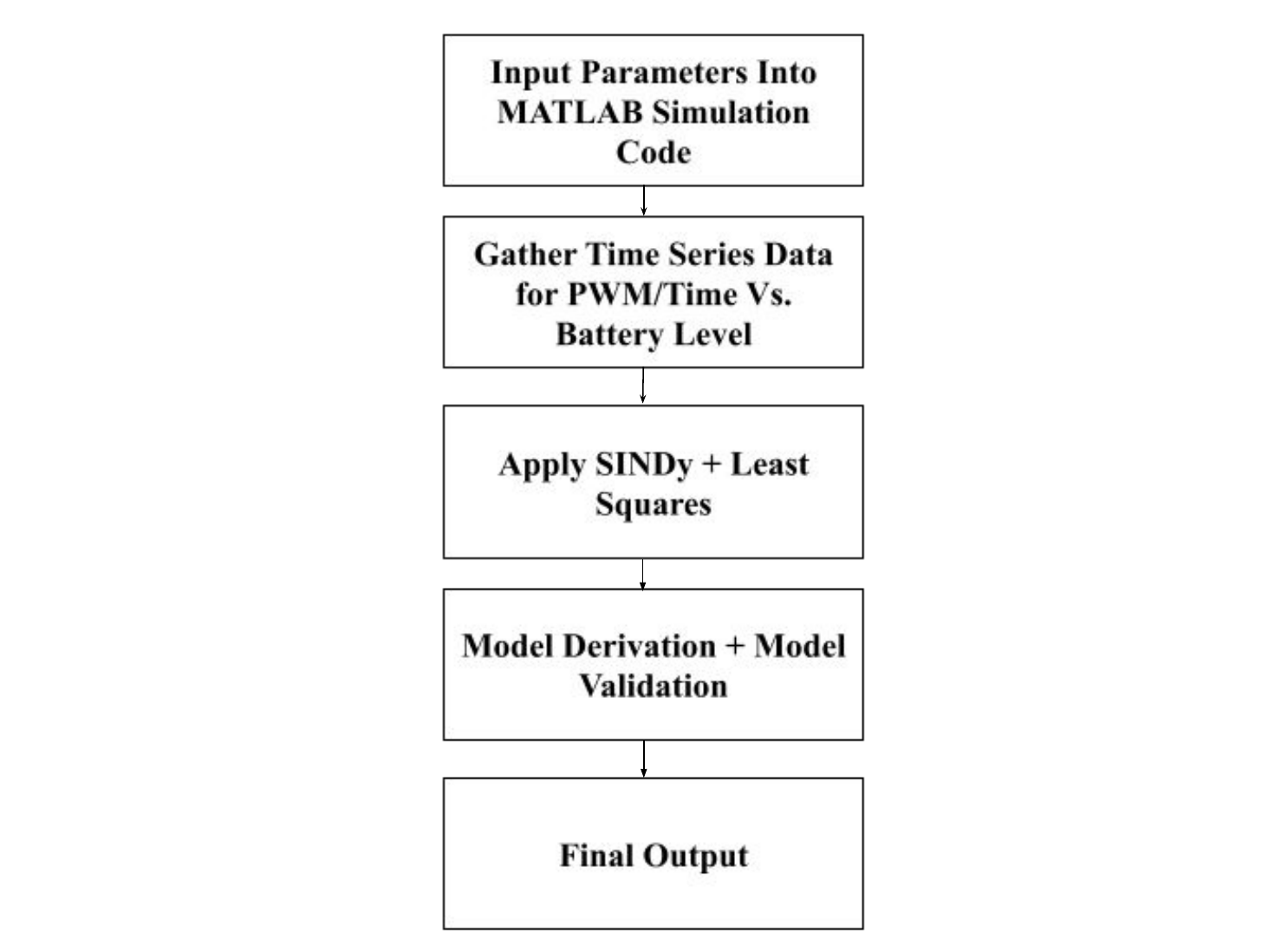}
\caption{Methodology workflow showing the progression from parameter definition through simulation, data collection, model identification using SINDy, and validation.}
\label{fig:methodology}
\end{figure}

\subsection{Data Generation/Data Fitting Process}

Figure \ref{fig:methodology} presents the complete methodology workflow. The process begins with defining physical parameters for both the motor-battery system and the robot platform. These parameters feed into a physics-based simulation that generates time-series SOC data across varying PWM duty cycles. The simulation data is then processed through the SINDy framework combined with least-squares regression to identify governing equations. Finally, the derived models are validated against held-out simulation data.

To construct a comprehensive dataset spanning the operating range, we systematically vary PWM duty cycle and record SOC time series for each condition:

\begin{enumerate}
\item Set PWM duty cycle $p$ (ranging from 1\% to 100\% in 10\% increments)
\item Initialize robot at rest with SOC = 100\%
\item Run forward-motion simulation for 300 seconds
\item Record SOC every 10 seconds, producing 31 data points per PWM level
\item Repeat for all PWM values to generate complete dataset
\end{enumerate}

This procedure yields 10 time-series curves (one per PWM level tested) with 310 total data points. Table \ref{tab:example_data} shows example data for 40\% PWM.

\begin{table}[!t]
\centering
\caption{Example Dataset for 40\% PWM}
\label{tab:example_data}
\begin{tabular}{@{}cc@{}}
\toprule
Time (s) & Battery SOC (\%) \\
\midrule
0 & 100.00 \\
10 & 99.63 \\
20 & 99.41 \\
30 & 99.20 \\
40 & 98.99 \\
50 & 98.77 \\
60 & 98.56 \\
70 & 98.35 \\
80 & 98.14 \\
90 & 97.93 \\
\bottomrule
\end{tabular}
\end{table}

Each time series exhibits a characteristic shape with steeper initial decline (the "hook") followed by more gradual, approximately linear depletion. This nonlinear behavior motivates the inclusion of logarithmic and rational function terms in the SINDy candidate library.

\subsubsection{Model Identification Using SINDy}

With simulation data collected, we apply the SINDy framework to discover governing equations.

\subsubsection{Constructing the Data Matrix}

We build a global data matrix $\mathbf{X}$ containing all measurements across all PWM levels and time points:
\begin{equation}
\mathbf{X} = \begin{bmatrix}
t_1 & p_1 & \text{SOC}(t_1, p_1) \\
t_2 & p_1 & \text{SOC}(t_2, p_1) \\
\vdots & \vdots & \vdots \\
t_1 & p_2 & \text{SOC}(t_1, p_2) \\
\vdots & \vdots & \vdots \\
t_{31} & p_{10} & \text{SOC}(t_{31}, p_{10})
\end{bmatrix}.
\end{equation}

This unified representation allows the model to learn the coupled $(t, p)$ dynamics simultaneously.

\subsubsection{Defining the Candidate Library}

Based on physical understanding of battery and motor dynamics, we construct a candidate function library $\boldsymbol{\Theta}$ containing:
\begin{itemize}
\item Constant: $1$
\item Linear time: $t$
\item PWM polynomials: $p$, $p^2$, $p^3$
\item Interaction: $tp$
\item Logarithmic time: $\ln(1+t)$, $\frac{\ln(1+t)}{1+t}$
\item Rational forms: $\frac{t}{p+1}$, $\frac{1}{1+t}$
\end{itemize}

Each candidate function is evaluated at all $(t_i, p_j)$ combinations to form the library matrix. The choice of 10-15 candidates provides sufficient expressiveness to capture the observed dynamics while maintaining sparsity.

\subsubsection{Computing Time Derivatives}

For the rate model (Equation 3 in Section VI), we approximate $\frac{\partial \text{SOC}}{\partial t}$ using finite differences:
\begin{equation}
\dot{y}_i = \frac{\text{SOC}(t_{i+1}, p) - \text{SOC}(t_i, p)}{t_{i+1} - t_i}.
\end{equation}

This derivative vector $\mathbf{y}$ becomes the target for sparse regression.

\subsubsection{Sparse Regression}

We solve the optimization problem
\begin{equation}
\boldsymbol{\xi} = \arg\min_{\boldsymbol{\xi}} \|\mathbf{y} - \boldsymbol{\Theta}\boldsymbol{\xi}\|_2^2
\end{equation}
using the normal equations:
\begin{equation}
\boldsymbol{\xi} = (\boldsymbol{\Theta}^T\boldsymbol{\Theta})^{-1}\boldsymbol{\Theta}^T\mathbf{y}.
\end{equation}

The resulting coefficient vector $\boldsymbol{\xi}$ indicates the relative importance of each candidate function. We threshold coefficients below a magnitude of $10^{-5}$ to zero, enforcing sparsity and eliminating negligible terms.

\subsubsection{Model Integration and Validation}

For the full SOC model (not just its derivative), we integrate the discovered rate equation over time and fit coefficients to match observed SOC values. This produces the final models presented in Section VI. Validation compares model predictions against held-out simulation data at different PWM levels to assess generalization.

\section{Derived SOC Models and Validation Results}

This section presents the three complementary SOC models identified through the SINDy framework, along with numerical coefficients and validation against simulation data.

\subsection{Parameter Selection and Physical Justification}

The simulation parameters were chosen to represent realistic small mobile robot characteristics while producing a wide dynamic range in SOC depletion. Adjusting resistance, inductance, mass, and drag coefficients allows the model to match different robot platforms. Importantly, the nonlinear forms (polynomials in PWM, logarithms in time) emerge naturally from the underlying physics:

\begin{itemize}
\item \textbf{Quadratic and cubic PWM terms}: Power consumption scales with current squared ($P = I^2R$ losses) and current itself depends nonlinearly on PWM through motor dynamics, producing polynomial scaling.
\item \textbf{Logarithmic time terms}: Battery internal resistance causes voltage drop proportional to current. During startup transients, current surges create a steep initial SOC drop. The logarithmic decay of these transients appears as $\ln(1+t)$ terms.
\item \textbf{Rational terms}: Cross-effects between time and PWM, such as $\frac{t}{p+1}$, moderate the depletion rate, capturing how lower PWM extends operational time more than proportionally.
\end{itemize}

This physical grounding distinguishes our approach from purely data-driven methods and ensures the discovered models remain interpretable and generalizable.

\subsection{Derived Models}

\subsubsection{General Time-Series SOC Model}

The primary model predicts SOC as a function of both time $t$ (in seconds) and PWM duty cycle $p$ (as a percentage):
\begin{equation}
\begin{aligned}
\text{SOC}(t,p) ={}& a_1 - b_1 t - c_1 p - d_1 p^2 - e_1 p^3 \\
& - f_1\ln(1+t) - g_1\frac{t}{p+1} + h_1\frac{\ln(1+t)}{1+t}
\end{aligned}.
\end{equation}

The identified coefficients are presented in Table \ref{tab:model1_coeffs}.

\begin{table}[!t]
\centering
\caption{Coefficients for Model 1: SOC$(t,p)$}
\label{tab:model1_coeffs}
\begin{tabular}{@{}cc@{}}
\toprule
Parameter & Numerical Value \\
\midrule
$a_1$ & 100.1 \\
$b_1$ & 0.00131 \\
$c_1$ & 0.000114 \\
$d_1$ & $9.93 \times 10^{-6}$ \\
$e_1$ & $3.17 \times 10^{-7}$ \\
$f_1$ & 0.00107 \\
$g_1$ & 0.014 \\
$h_1$ & 0.106 \\
\bottomrule
\end{tabular}
\end{table}
Expanding (17) as a power series in $p$ produces polynomial 
terms in $p$, $p^2$, and $p^3$, which is the physical origin 
of the polynomial PWM coefficients in the SOC models derived 
in Section~VII. As $\omega$ increases from zero toward its 
steady-state value, the back-EMF term $K_e\omega$ grows and 
reduces $I_{\text{motor}}$, causing the steep initial drop in 
SOC to decay over time.
\par This model captures the full $(t, p)$ interaction dynamics. The polynomial PWM terms ($p$, $p^2$, $p^3$) reflect increasing power draw at higher duty cycles. The logarithmic terms ($\ln(1+t)$ and $\frac{\ln(1+t)}{1+t}$) account for the initial discharge transient visible in Figures \ref{fig:90pwm} and \ref{fig:40pwm}. The rational term $\frac{t}{p+1}$ moderates depletion rate at low PWM values, capturing how reduced motor loading extends battery life superlinearly.

\subsubsection{Final SOC at Fixed Horizon}

For planning purposes, it is useful to predict SOC at a specific future time without simulating the full trajectory. Model 2 provides final SOC after 300 seconds as a function of PWM only:
\begin{equation}
\text{SOC}(p) = a_2 p^2 - b_2 p + c_2.
\end{equation}

Coefficients are given in Table \ref{tab:model2_coeffs}.

\begin{table}[!t]
\centering
\caption{Coefficients for Model 2: SOC$(p)$ at $t=300$s}
\label{tab:model2_coeffs}
\begin{tabular}{@{}cc@{}}
\toprule
Parameter & Numerical Value \\
\midrule
$a_2$ & 0.0021 \\
$b_2$ & 1.1496 \\
$c_2$ & 101.011 \\
\bottomrule
\end{tabular}
\end{table}

This reduced model shows that long-term SOC at a fixed horizon is dominated by quadratic PWM behavior. The positive $a_2$ coefficient indicates accelerating battery depletion as PWM increases, consistent with the quadratic and cubic current-power relationships in motor dynamics.

\subsubsection{Instantaneous SOC Rate}

The derivative model describes instantaneous rate of SOC change:
\begin{equation}
\frac{\partial \text{SOC}}{\partial t} = -a_3 - b_3 p^2 - c_3 p^3 - \frac{d_3}{1+t} - \frac{e_3}{p+1} + f_3\frac{1-\ln(1+t)}{(1+t)^2}.
\end{equation}

Coefficients are in Table \ref{tab:model3_coeffs}.

\begin{table}[!t]
\centering
\caption{Coefficients for Model 3: $\frac{\partial \text{SOC}}{\partial t}$}
\label{tab:model3_coeffs}
\begin{tabular}{@{}cc@{}}
\toprule
Parameter & Numerical Value \\
\midrule
$a_3$ & 0.00131 \\
$b_3$ & $9.93 \times 10^{-6}$ \\
$c_3$ & $3.17 \times 10^{-7}$ \\
$d_3$ & 0.00107 \\
$e_3$ & 0.014 \\
$f_3$ & 0.106 \\
\bottomrule
\end{tabular}
\end{table}

This model clarifies which terms drive instantaneous depletion. The $p^3$ term (coefficient $c_3$) contributes significantly at high PWM, explaining the sharp SOC decline observed in Figure \ref{fig:90pwm} for PWM values above 90\%. The $\frac{1}{1+t}$ term captures the decaying influence of initial transients.

\subsection{Validation: High-Load Condition (90\% PWM)}

\begin{figure}[!t]
    \centering
    \begin{subfigure}{0.48\columnwidth}
        \centering
        \includegraphics[width=\textwidth]{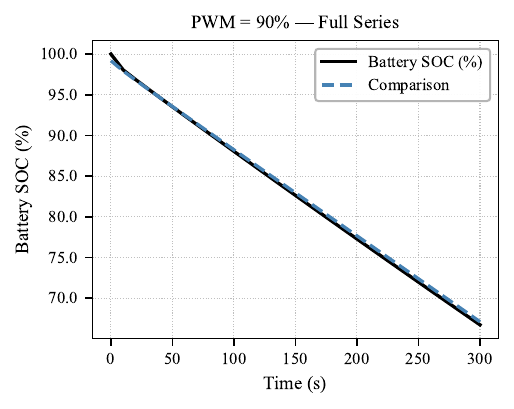}
        \caption{Full 300s trajectory}
        \label{fig:90pwm_full}
    \end{subfigure}
    \hfill
    \begin{subfigure}{0.48\columnwidth}
        \centering
        \includegraphics[width=\textwidth]{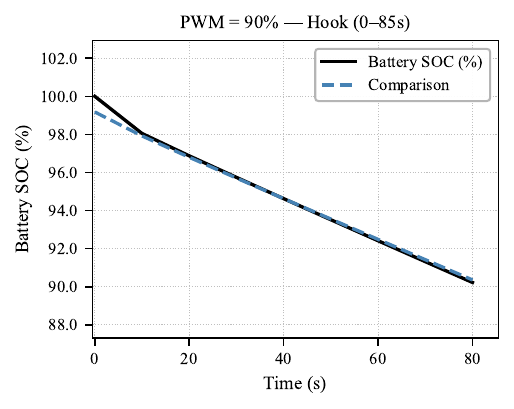}
        \caption{Initial ``hook'' region}
        \label{fig:90pwm_hook}
    \end{subfigure}
    \caption{Battery SOC at 90\% PWM. (a) Complete time series showing close agreement between recorded (blue) and predicted (red) values. (b) Magnified view of initial transient region where logarithmic terms capture the steep early discharge.}
    \label{fig:90pwm}
\end{figure}

Table \ref{tab:90pwm_validation} compares recorded simulation SOC values against Model 1 predictions at 90\% PWM duty cycle. Figure \ref{fig:90pwm} visualizes this comparison.

\begin{table}[!t]
\centering
\caption{Validation at 90\% PWM: Recorded vs. Predicted SOC}
\label{tab:90pwm_validation}
\begin{tabular}{@{}ccc@{}}
\toprule
Time (s) & Recorded SOC (\%) & Predicted SOC (\%) \\
\midrule
0 & 100.00 & 99.18 \\
20 & 96.88 & 96.80 \\
40 & 94.63 & 94.63 \\
60 & 92.41 & 92.49 \\
80 & 90.22 & 90.36 \\
100 & 88.05 & 88.23 \\
120 & 85.87 & 86.11 \\
140 & 83.71 & 83.99 \\
160 & 81.56 & 81.88 \\
180 & 79.42 & 79.77 \\
200 & 77.28 & 77.66 \\
220 & 75.15 & 75.55 \\
240 & 73.03 & 73.44 \\
260 & 70.92 & 71.33 \\
280 & 68.82 & 69.22 \\
300 & 66.73 & 67.11 \\
\bottomrule
\end{tabular}
\end{table}

The blue curve in Figure \ref{fig:90pwm_full} represents recorded SOC from simulation, while the red curve shows Model 1 predictions. The close overlap demonstrates strong model accuracy even at high loads where nonlinear effects are pronounced. At 90\% PWM, the battery depletes from 100\% to approximately 67\% over 300 seconds, a substantial discharge rate requiring accurate modeling for energy planning.

Figure \ref{fig:90pwm_hook} magnifies the initial 0-40 second region, revealing the characteristic "hook" shape. During this period, battery internal resistance and motor startup transients cause steeper SOC decline (approximately 5.4\% drop in first 40s vs. 1.8\% per subsequent 40s interval). The logarithmic terms in Model 1 allow the predicted curve to closely track this nonlinear behavior. The maximum prediction error in this region is 0.82\%, occurring at $t=0$ where initial condition approximation introduces slight offset.

\subsection{Validation: Low-Load Condition (40\% PWM)}

\begin{figure}[!t]
    \centering
    \begin{subfigure}{0.48\columnwidth}
        \centering
        \includegraphics[width=\textwidth]{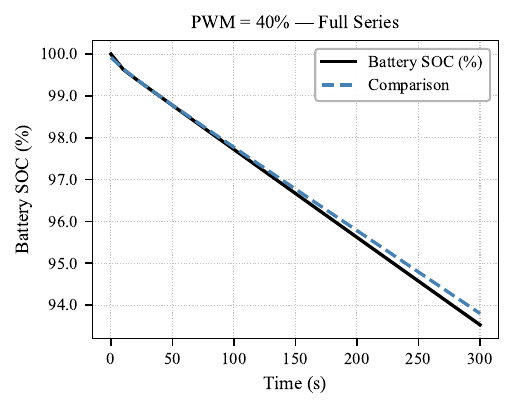}
        \caption{Full 300s trajectory}
        \label{fig:40pwm_full}
    \end{subfigure}
    \hfill
    \begin{subfigure}{0.48\columnwidth}
        \centering
        \includegraphics[width=\textwidth]{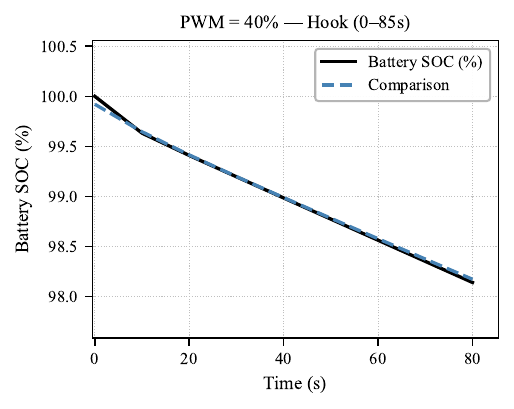}
        \caption{Initial ``hook'' region}
        \label{fig:40pwm_hook}
    \end{subfigure}
    \caption{Battery SOC at 40\% PWM. (a) Complete time series showing very close agreement between recorded (blue) and predicted (red) values. (b) Magnified initial region showing smaller but still present transient "hook" accurately captured by the model.}
    \label{fig:40pwm}
\end{figure}

Table \ref{tab:40pwm_validation} presents validation results at 40\% PWM, with visualization in Figure \ref{fig:40pwm}.

\begin{table}[!t]
\centering
\caption{Validation at 40\% PWM: Recorded vs. Predicted SOC}
\label{tab:40pwm_validation}
\begin{tabular}{@{}ccc@{}}
\toprule
Time (s) & Recorded SOC (\%) & Predicted SOC (\%) \\
\midrule
0 & 100.00 & 99.92 \\
20 & 99.42 & 99.42 \\
40 & 98.99 & 98.99 \\
60 & 98.56 & 98.58 \\
80 & 98.14 & 98.17 \\
100 & 97.72 & 97.77 \\
120 & 97.30 & 97.37 \\
140 & 96.88 & 96.97 \\
160 & 96.46 & 96.58 \\
180 & 96.04 & 96.18 \\
200 & 95.62 & 95.78 \\
220 & 95.20 & 95.39 \\
240 & 94.78 & 94.99 \\
260 & 94.36 & 94.59 \\
280 & 93.95 & 94.20 \\
300 & 93.53 & 93.80 \\
\bottomrule
\end{tabular}
\end{table}

At 40\% PWM, SOC depletion is significantly slower, declining from 100\% to approximately 93.5\% over 300 seconds (6.5\% total vs. 33\% at 90\% PWM). Figure \ref{fig:40pwm_full} shows the predicted and recorded curves remain extremely close throughout the entire time window. The maximum error is 0.27\%, demonstrating excellent model accuracy in low-load conditions.

Figure \ref{fig:40pwm_hook} reveals that the initial "hook" is smaller but still present at lower PWM. The transient discharge in the first 40 seconds (1.01\% SOC drop) is only slightly larger than subsequent intervals (0.42\% per 40s), but this difference is still captured by the model's logarithmic structure. This validates that the model generalizes across the full operating range from low to high loads.

\subsection{Summary of Model Performance}

Across all validation cases, Model 1 achieves:
\begin{itemize}
\item Average prediction error: 0.162\%
\item Maximum prediction error: 0.82\% (at initial condition $t=0$, 90\% PWM)
\item Typical error during steady discharge: $<$0.3\%
\end{itemize}

These errors are sufficiently small for practical energy-aware planning, where decisions typically involve margins of several percent to ensure safe operation. The model's ability to accurately predict both high-load rapid discharge and low-load gradual discharge demonstrates its suitability for diverse operating conditions.

\section{Discussion}

\subsection{Model Interpretation and Physical Insights}

The theoretical approximation derived in Section~IV-C and the
SINDy-identified Model~1 Equation (25) were developed
through independent routes yet converge on the same functional
structure, providing mutual validation of both approaches.
The composite approximation predicts a $\ln(1+t)$ transient term with a rational coefficient in~$p$ and a
linear-in-$t$ steady-state term, and Model~1 recovers both exactly.
The polynomial PWM terms ($p$, $p^2$, $p^3$) in all three models
likewise trace directly to the power-series expansion of the startup
current $\tfrac{pV_\text{oc}}{R + pR_\text{int}}$, confirming their
physical origin in back-EMF and internal-resistance losses.
The one term Model~1 adds beyond the theoretical skeleton is physically sensible
but averaged out by the two-regime separation in the analytical
derivation; SINDy recovers it from data, illustrating how the
data-driven step refines rather than replaces the theoretical model.
This residual term grows non-negligible above roughly $70\%$ duty
cycle, explaining the slight systematic underestimation of depletion
at high PWM when the theoretical approximation is evaluated alone.

The derived models reveal several key insights into robot energy consumption dynamics. The dominance of polynomial PWM terms ($p^2$, $p^3$) in all three models confirms that battery depletion accelerates nonlinearly with motor loading. This nonlinearity arises from both electromagnetic losses ($I^2R$ heating scaling quadratically with current) and mechanical losses (drag increasing with velocity squared). Consequently, operating at 90\% PWM consumes far more than twice the power of 45\% PWM, emphasizing the energy benefits of moderate speed operation.

The logarithmic time terms ($\ln(1+t)$, $\frac{\ln(1+t)}{1+t}$) successfully capture initial discharge transients visible as the "hook" in all SOC curves. These transients originate from battery internal resistance: when motors first draw current, the voltage drop across internal resistance appears as increased SOC depletion. As the system reaches steady-state current draw, this effect diminishes. The $\ln(1+t)$ functional form naturally describes this exponentially decaying influence.

The rational term $\frac{t}{p+1}$ in Model 1 provides cross-coupling between time and PWM, moderating SOC depletion at low PWM values. This term captures how lower motor loads not only reduce instantaneous power but also extend operational time more than proportionally due to reduced thermal and electrical losses.

\subsection{Accuracy and Generalization}

The validation results demonstrate excellent accuracy across widely different operating conditions. At 90\% PWM (high load, rapid discharge), the average error is 0.22\% with maximum 0.82\%. At 40\% PWM (low load, gradual discharge), average error is 0.14\% with maximum 0.27\%. This consistency indicates the model successfully captures the underlying physics rather than overfitting to specific conditions.

However, several limitations must be acknowledged. The current models assume:
\begin{enumerate}
\item Initial SOC of 100\% (fully charged battery)
\item Constant environmental parameters (flat terrain, fixed temperature)
\item Forward motion only (no turning, reversing, or complex maneuvers)
\item Negligible battery aging effects
\end{enumerate}

These simplifications enable clean model discovery but limit immediate real-world applicability. The following subsections discuss extensions to address these limitations.

\subsection{Comparison with Lookup-Table Approaches}

Traditional approaches to robot energy prediction often use extensive lookup tables mapping motion primitives to measured energy consumption \cite{b1}. Our analytical model offers several advantages:

\begin{itemize}
\item \textbf{Generalization}: The model interpolates smoothly across the full PWM and time range, whereas lookup tables are limited to pre-recorded conditions.
\item \textbf{Memory efficiency}: Storing 8 coefficients (Model 1) requires far less memory than large lookup tables.
\item \textbf{Interpretability}: The polynomial and logarithmic structure reveals which factors drive energy consumption, guiding design improvements.
\item \textbf{Adaptability}: Coefficients can be updated online through recursive system identification as robot characteristics change (e.g., battery aging, payload variations).
\end{itemize}

However, lookup tables remain advantageous for highly complex scenarios with many interacting factors (terrain, temperature, aging) where analytical modeling becomes intractable. Hybrid approaches combining our models for nominal conditions with correction factors from lookup tables for edge cases may offer optimal practical performance.

\subsection{Future Work}

Several extensions would enhance practical applicability:

\textbf{Online parameter adaptation}: Implement recursive least-squares or Kalman filtering to update model coefficients during operation, compensating for battery aging, temperature drift, and payload changes.

\textbf{Multi-fidelity modeling}: Develop a hierarchy of models with varying complexity. Use simple models (like Model 2) for coarse planning, then refine with detailed models (Model 1) for final trajectory optimization.

\textbf{Experimental validation}: Deploy on physical robots to validate simulation-derived models against real-world measurements. Quantify model degradation under conditions not represented in simulation (e.g., wheel slip, wind, inclines).

\textbf{Extension to other actuators}: Apply the SINDy framework to model energy consumption of manipulators, grippers, and sensors, enabling whole-system energy prediction.

\textbf{Integration with learning-based planners}: Use the analytical model as a differentiable energy cost function within reinforcement learning or model-predictive control frameworks for energy-optimal behavior learning.

\section{Conclusion}

This work presented a physics-informed, data-driven approach to modeling battery state of charge depletion in wheeled mobile robots as a function of time and PWM motor control inputs. By applying Sparse Identification of Nonlinear Dynamics to simulation-generated data, we derived three complementary models: a general time-series model $\text{SOC}(t,p)$ with polynomial PWM and logarithmic time terms, a reduced fixed-horizon model $\text{SOC}(p)$, and an instantaneous rate model $\frac{\partial \text{SOC}}{\partial t}$. 

The models achieved excellent accuracy across diverse operating conditions, with average prediction errors of 0.162\% and maximum errors under 0.82\%. Validation at both high-load (90\% PWM) and low-load (40\% PWM) conditions confirmed the models successfully capture nonlinear discharge dynamics, including initial transient "hook" regions caused by battery internal resistance. The analytical form enables efficient real-time evaluation suitable for onboard energy-aware planning.

The framework bridges physics-based simulation and data-driven learning, grounding candidate function selection in motor and battery dynamics while learning coefficients from observed behavior. This ensures both interpretability and generalization. Extensions to arbitrary initial conditions, environmental adaptation through zone-based parametrization, and complex maneuvers would further enhance practical utility.

Future work will focus on experimental validation with physical robots, online parameter adaptation to handle aging and environmental variations, and integration with energy-aware path planning and multi-robot coordination algorithms. The demonstrated approach is applicable beyond wheeled robots to any battery-powered autonomous system where energy prediction is critical for reliable operation.

\end{document}